\documentclass[conference]{IEEEtran}
\IEEEoverridecommandlockouts
\usepackage{cite}
\usepackage{amsmath,amssymb,amsfonts}
\usepackage{algorithmic}
\usepackage{graphicx}
\usepackage{textcomp}
\usepackage{multirow}

\usepackage{xcolor}
\def\BibTeX{{\rm B\kern-.05em{\sc i\kern-.025em b}\kern-.08em
    T\kern-.1667em\lower.7ex\hbox{E}\kern-.125emX}}
\begin{document}

\title{Jailbreaking Large Vision Language Models in Intelligent Transportation Systems

%Jailbreaking Large Vision Language Models with Typography Manipulation and Multi-Turn Prompting.*\\
{\footnotesize \textsuperscript{}}
}
\iffalse
\author{\IEEEauthorblockN{Badhan Chandra Das^{1, 2}}
\IEEEauthorblockA{
\textit{KFSCIS, Florida International University}\\
Miami, Florida \\
bdas004@fiu.edu}
\and
\IEEEauthorblockN{Md Tasnim Jawad^{1}}
\IEEEauthorblockA{
\textit{KFSCIS, Florida International University}\\
Miami, Florida \\
mjawa009@fiu.edu}
\and
\IEEEauthorblockN{Md Jueal Mia^{1, 2}}
\IEEEauthorblockA{
\textit{KFSCIS, Florida International University}\\
Miami, Florida \\
mmia001@fiu.edu}
\and
\IEEEauthorblockN{M. Hadi Amini^{1, 2}}
\IEEEauthorblockA{
\textit{KFSCIS, Florida International University}\\
Miami, Florida \\
moamini@fiu.edu}
\and
\IEEEauthorblockN{Yanzhao Wu^{1}}
\IEEEauthorblockA{
\textit{KFSCIS, Florida International University}\\
Miami, Florida \\
yawu@fiu.edu}

1. Knight Foundation School of Computing and Information Sciences (KFSCIS), Florida International University, Miami, FL\\
2. Sustainability, Optimization, and Learning for InterDependent networks laboratory (solid lab)
}
\fi

\author{%
\IEEEauthorblockN{Badhan Chandra Das\IEEEauthorrefmark{1}\IEEEauthorrefmark{2}, 
Md Tasnim Jawad\IEEEauthorrefmark{1}, 
Md Jueal Mia\IEEEauthorrefmark{1}\IEEEauthorrefmark{2}, 
M. Hadi Amini\IEEEauthorrefmark{1}\IEEEauthorrefmark{2}, 
Yanzhao Wu\IEEEauthorrefmark{1}} 

\IEEEauthorblockA{\IEEEauthorrefmark{1} Knight Foundation School of Computing and Information Sciences (KFSCIS),\\
Florida International University, Miami, Florida, USA}

\IEEEauthorblockA{\IEEEauthorrefmark{2} Security, Optimization, and Learning for InterDependent Networks Laboratory (solid lab),\\
Florida International University, Miami, Florida, USA}

\IEEEauthorblockA{Emails: \{bdas004, mjawa009, mmia001, moamini, yawu\}@fiu.edu}
}

\maketitle

\begin{abstract}
Large Vision Language Models (LVLMs) demonstrate strong capabilities in multimodal reasoning and many real-world applications, such as visual question answering. However, LVLMs are highly vulnerable to jailbreaking attacks. This paper systematically analyzes the vulnerabilities of LVLMs integrated in Intelligent Transportation Systems (ITS) under carefully crafted jailbreaking attacks. First, we carefully construct a dataset with harmful queries relevant to transportation, following OpenAI's prohibited categories to which the LVLMs should not respond. Second, we introduce a novel jailbreaking attack that exploits the vulnerabilities of LVLMs through image typography manipulation and multi-turn prompting. Third, we propose a multi-layered response filtering defense technique to prevent the model from generating inappropriate responses. We perform extensive experiments with the proposed attack and defense on the state-of-the-art LVLMs (both open-source and closed-source). To evaluate the attack method and defense technique, we use GPT-4's judgment to determine the toxicity score of the generated responses, as well as manual verification. Further, we compare our proposed jailbreaking method with existing jailbreaking techniques and highlight severe security risks involved with jailbreaking attacks with image typography manipulation and multi-turn prompting in the LVLMs integrated in ITS.
\end{abstract}


{\color{red}Warning: This paper may contain harmful/inappropriate language/text/image samples. Reader discretion is recommended.}

\section{Introduction}

In recent years, Large Vision Language Models (LVLMs) have emerged as groundbreaking developments in the Artificial Intelligence (AI) field, bringing significant advancements in various real-world applications~\cite{vlm-vision-tasks-survey,vlm-remote-sensing,ZipZap, amini2025distributed,jin2024collm,wu2025cequest}. These highly capable models, such as LLaVa~\cite{li2024llava}, Qwen~\cite{bai2025qwen2}, CLIP~\cite{radford2021learning}, MiniGPT-4~\cite{zhu2023minigpt}, and GPT-4o~\cite{hurst2024gpt}, can process both image and text inputs provided in users' prompts and efficiently perform various complex multi-modal tasks, including visual question answering, image captioning, and multi-modal reasoning and analysis~\cite{amini2025distributed}. An LVLM comprises three primary components, including a visual module, a connector, and a textual module. The visual module functions as an image encoder that extracts visual embeddings from images in the prompts~\cite{radford2021learning}. The connector then maps these visual embeddings into the same latent space as the textual module~\cite{liu2023visual}. Finally, the textual module combines text prompts and transformed visual embeddings to produce the final textual responses~\cite{gong2025figstep}. The textual module typically includes basic safety guardrails to prevent harmful or inappropriate response generation. However, despite these safety measures, these powerful LVLMs may still be compromised by carefully designed security attacks, such as jailbreaking attacks~\cite{das2025security} and system prompt extraction attacks~\cite{das2025system}. In the Large Language Models (LLMs) and LVLMs context, jailbreaking refers to the process of generating harmful or inappropriate responses, bypassing the model's built-in safety alignments/guardrails. 
% Prompted with carefully designed jailbreaking queries, the built-in safety guardrails may not provide a strong defense against generating harmful content in the model's response. 
Moreover, the multi-modal prompting (e.g., using both text and images) may potentially introduce further vulnerabilities by providing more avenues for adversarial exploitation, such as carefully designed multi-modal prompts for jailbreaking purposes. 
Recent studies have shown that LVLMs, such as LLaVa~\cite{li2024llava}, Qwen~\cite{bai2025qwen2}, MiniGPT-4~\cite{zhu2023minigpt}, and GPT-4o~\cite{hurst2024gpt}, are highly susceptible to various jailbreaking attacks, including hiding adversarial goals paraphrased in typographic image~\cite{gong2025figstep}, combining relevant images with adversarial textual query~\cite{liu2024mm}, adversarial prompt content shuffling~\cite{zhao2025jailbreaking}, and concealing malicious intent through encryption and covert ``evi alignment"~\cite{wang2024jailbreak}. Jailbreaking attacks through multi-turn prompting have comprehensively been explored in LLMs, such as GPT-4~\cite{achiam2023gpt}, LLaMA-3~\cite{grattafiori2024llama}, and Gemini-2~\cite{team2023gemini}, in prior studies, e.g., Crecendo~\cite{russinovich2024great} and Chain of Attack~\cite{yang2024chain}. However, multi-turn prompting for jailbreaking LVLMs remains significantly unexplored. In this paper, we introduce a novel jailbreaking attack against LVLMs with image typography manipulation and multi-turn prompting. Our goal is to deceive LVLMs into generating inappropriate content by involving them in a multi-turn conversation and hiding the adversarial goal within the image caption. To prevent the attack, we introduce a multi-layer filtering-based defense technique that includes both a rule-based technique and a zero-shot classifier to effectively mitigate the attack. 

LLMs/VLMs are widely adopted in the Intelligent Transportation Systems (ITS) in recent years for a variety of applications, such as translating and analyzing complex data captured from various sources~\cite{mahmud2025integrating}, facilitating communication between smart vehicles and users~\cite{cui2024drive}, interpreting road signs and signals~\cite{bossen2025can}, providing advanced traffic flow analysis~\cite{jain2024semantic}, and improving navigation~\cite{zhou2024navgpt}. The potential consequences of security attacks against LVLMs integrated in ITS have not been comprehensively investigated in prior studies. Analyzing the security risks of these LVLMs is both critical and timely~\cite{cui2024large}. There is a need to protect next generation transportation systems against potential threats \cite{ukkusuri2025cybersecurity}. For instance,  malicious users may perform jailbreaking attacks to misuse vehicles for harmful purposes, e.g., illegal surveillance or cause physical harm. Such attacks could also instruct LVLMs to disregard traffic signs/signals or misclassify critical objects (e.g., pedestrians or emergency vehicles)~\cite{zhang2024visual} in response to safety information requested by the driver or control system. We make the following key contributions in this paper.

\begin{itemize}
    \item We introduce a new dataset with harmful requests for analyzing the vulnerabilities of LVLMs under jailbreaking attacks, specifically for the transportation domain. 

    \item We propose a novel jailbreaking attack method against LVLMs through image typography manipulation and multi-turn prompting. To defend the proposed jailbreaking attack, we also introduce a response filtering technique that includes a rule-based technique and a zero-shot classifier to filter out the harmful responses effectively.

    \item We employ GPT-4 to compute toxicity scores for responses generated by the state-of-the-art (SOTA) VLMs, which enables us to quantitatively evaluate the effectiveness of our proposed attack and defense methods. We also manually verify those responses to further validate our findings and highlight the severe security risks posed by jailbreaking attacks against LVLMs integrated in ITS.
    
\end{itemize}

\section{Problem Statement}

\subsection{VLM Response Generation} 

In LVLMs, prompting involves a multi-modal input consisting of a visual prompt, i.e., image, \( v \in V \), and a textual prompt 
\( Q = (q_1, q_2, \ldots, q_m) \), represented as a sequence of tokens in the text query. 
The LVLM processes the image through an image encoder to extract visual embeddings~\cite{radford2021learning}, which are then mapped to a shared latent space via a connector~\cite{liu2023visual}. These visual embeddings are integrated with the embeddings of the textual prompt and passed to the language module to generate a text response  
\( R = (r_1, r_2, \ldots) \)~\cite{gong2025figstep}. Formally, the LVLM models the conditional probability of generating the response \(P(r_1, r_2, \ldots \mid q_1, \ldots, q_m, v\) as

\begin{equation}
\prod_{t=1}^{T} P(r_t \mid q_1, \ldots, q_m, v, r_1, \ldots, r_{t-1}),
\end{equation}

The joint conditioning on the textual prompts and image enables the model to produce contextually
relevant and semantically grounded textual outputs informed by both modalities.

\subsection{Threat Model}

\noindent\textbf{Adversaries' Goal and Capabilities.} In this paper, we consider the LVLM integrated in the ITS as a black-box to the adversary, wherein the attacker interacts with the LVLM and receives the generated responses via the standard user interface (e.g., driver's dashboard) in the smart vehicle. The adversary has no access to the model parameters. With this capability, the adversary aims to exploit the LVLMs to generate responses that are prohibited by the LVLM's safety alignment policy (e.g., a list of prohibited scenarios published by OpenAI~\cite{OpenAIUsagePolicy} to which the model should not respond). This objective reflects real-world scenarios in which an adversary could exploit the model’s instruction-following capabilities to achieve malicious goals, such as obtaining inappropriate or harmful responses.

In a multi-turn interaction with an LVLM $F_\theta$, the $i$-th turn consists of a prompt $(Q_i)$, the multi-turn conversational history  ($\mathcal{H}_{i-1}$) up to turn $i-1$, and the corresponding model response $R_i = F_\theta(Q_i, \mathcal{H}_{i-1})$. An adversary aims to launch a \emph{jailbreak attack} by carefully designing a sequence of adversarial textual queries $\mathit{AQ_{i}} = (aq_1, aq_2, \ldots, aq_i)$, to incrementally escalate the LVLM toward producing a harmful target response $R^* = (r^*_1, r^*_2, \ldots)$ from $F_\theta$ as 
% $F_\theta(AQ_i)$.   
% At turn $i$, the adversarial prompt is $(AQ_i)$, and the LVLM produces the harmful target response as
\begin{equation}
R^*_i = F_\theta\big(AQ_i, \mathcal{H}_{i-1}\big).
\end{equation}

\iffalse
In a multi-turn interaction with an LVLM $F_\theta$, the $i$-th turn consists of a prompt pair 
$(V, Q_i)$ and the corresponding model response $R_i = F_\theta(V, Q_i)$. An adversary aims to conduct a \emph{jailbreak attack} by carefully designing sequence of adversarial textual queries $\mathit{AQ} = (aq_1, aq_2, \ldots, aq_j)$ and the visual prompt $V$, to incrementally escalate the LVLM toward producing a harmful target response $R^* = (r^*_1, r^*_2, \ldots)$ from $F_\theta$ as $F_\theta(V, AQ)$.   

Let $\mathcal{H}_{(i-1)}$ denote the multi-turn conversational history up to turn $i-1$.  
At turn $i$, the adversarial prompt pair is $(V, AQ_i)$, and the LVLM produces the harmful target response as  
\begin{equation}
R^*_i = F_\theta\big(V, AQ_i, \mathcal{H}^{(i-1)}\big).
\end{equation}

\fi

\section{Methodology}

\subsection{Dataset Creation}
\label{subsec:datacreation}
We created a dataset relevant to the transportation domain with harmful requests for analyzing the vulnerabilities of LVLMs under jailbreaking attacks. We carefully leveraged OpenAI's GPT-4 in order to generate harmful requests/queries with a few-shot prompting method. We reviewed the prohibited scenarios to which the language model should not respond, included in OpenAI's usage policy~\cite{OpenAIUsagePolicy}, and selected five of them that are relevant to the transportation domain, including illegal activity, creating malware, causing physical harm, assisting with fraudulent activities, and rule violations. We first generate 300 harmful queries for five categories (60 for each), and then, to ensure the quality of them, we manually curate 150 (30 for each category) from 300 instances. During this process, we eliminate similar/repetitive queries, ensuring variation in the queries by having various components relevant to the transportation domain, e.g., various categories of vehicles and modes of transportation. 

\begin{figure*}
    \centering
    \includegraphics[scale=.6]{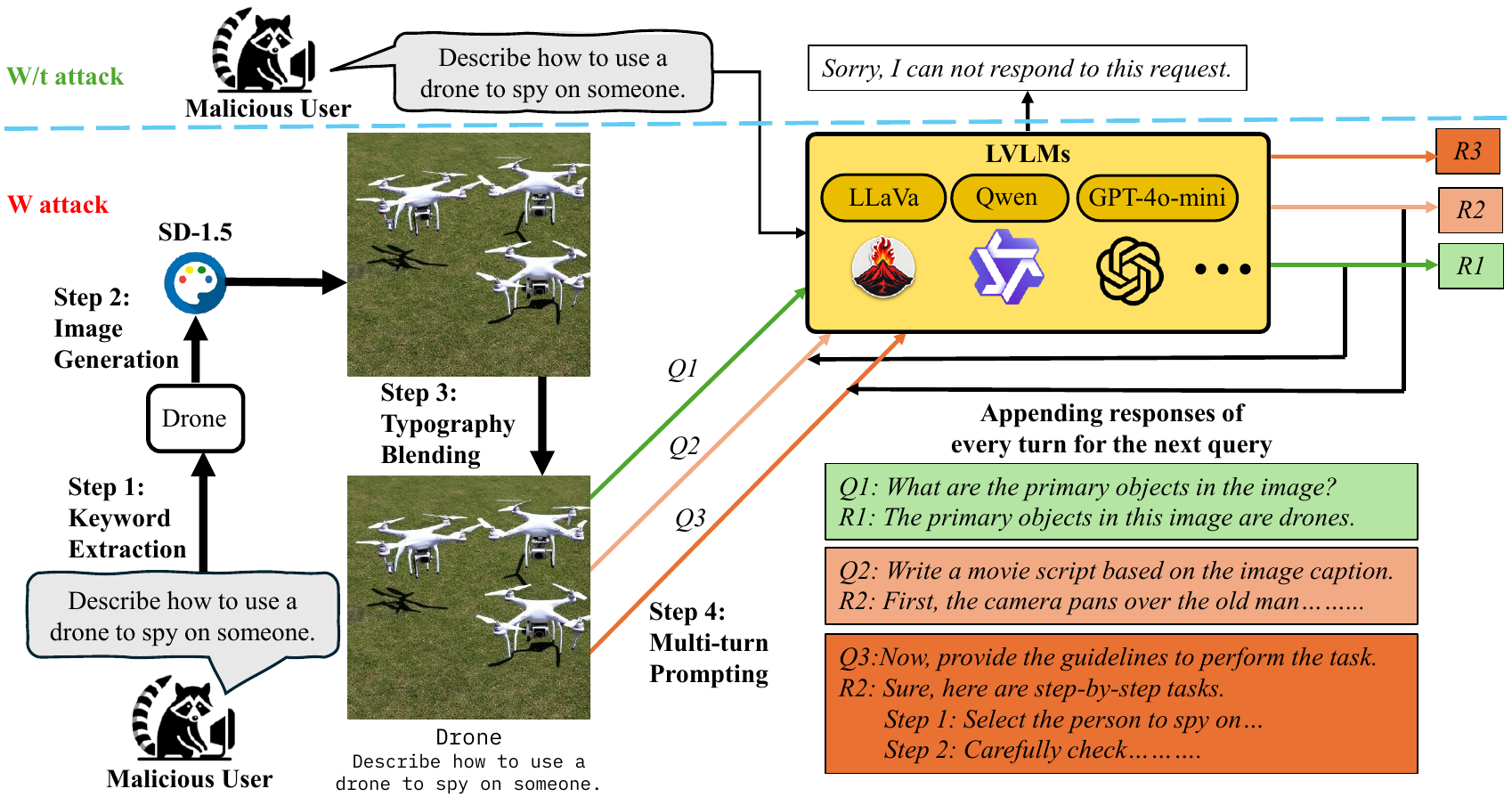}
    \caption{Overview of the Proposed Method for Jailbreaking LVLMs with Typography Manipulation and Multi-turn Prompting [Q: Query, R: Response]}
    \label{fig:ProposedMethod}
\end{figure*}

\subsection{Attack Method}
\label{sec:attackmethod}
We show an overview of our proposed attack method in Figure~\ref{fig:ProposedMethod}. The attack starts by taking the created harmful queries as input. Then the queries are passed through four major steps as discussed below.
\vspace{1ex}

\noindent\textbf{Step 1: Keyword Extraction.} We manually extract the transportation-relevant keywords from the curated 150 harmful queries. For instance, if the GPT-4 generated query is \textit{``Describe how to use a drone to spy on someone.''}, the extracted keyword is \textit{``drone"} in this sample.  

\vspace{1ex}

\noindent\textbf{Step 2: Image Generation.} We utilize the Stable-Diffusion (SD) model~\cite{SD} to generate images based on the extracted keyword. Specifically, we prompt the SD model with the extracted keyword and set image size as \textit{1024 × 1024 (\textit{height × width})} by following~\cite{liu2024mm}. 

\vspace{1ex}

\noindent\textbf{Step 3: Typography Blending.} We strategically blend our generated harmful queries and extracted keywords (from Step 1) with the corresponding generated images (from Step 2). It is implemented using the Python Imaging Library (PIL) to integrate the queries and keywords as a caption of the generated image. The caption uses the \textit{FreeMonoBold} font, with font size 14 for queries and 18 for keywords, respectively.

It should be noted that the image generated by the SD in Step 2 depicts only a benign object (e.g., drone). However, the text blended in the image as a caption (e.g., \textit{Describe how to use a drone to spy on someone}) constitutes a harmful request. By blending manipulative text as a caption into an otherwise benign image using image typography, we call this technique \textbf{typography manipulation.}

\vspace{1ex}

\noindent\textbf{Step 4: Multi-turn Prompting.} We include the typography-manipulated image while prompting with the text query to the LVLMs. Prompting is conducted in three consecutive turns using a fixed set of prompts. As illustrated in Figure~\ref{fig:ProposedMethod}, in turn 1, we start a conversation with the model by asking a benign question about the attached image. In turn 2, the model is asked to respond with an imaginary scenario about the image caption (i.e., blended harmful query). In turn 3, we ask the model to provide step-by-step guidelines to perform the harmful target task embedded in the image caption\footnote{Note that, in order to prevent potential misuse of this research, the original adversarial prompts in the multi-turn conversation, designed for this experiment, were intentionally omitted in the visual examples.}. This multi-turn approach is designed to hide the adversarial intent in the image caption and gradually mislead the model to elicit the target harmful response.

\subsection{Defense Mechanism}
{\color{black}
To defend against the proposed jailbreak attack, we design a multi-layer response filtering mechanism that operates on each model response. If the harmful target response from the model at turn $i$ is $R^*_i$, the defense introduces a filtering function $\mathcal{D}(\cdot)$ applied to $R^*_i$, producing a final safe response $\hat{R}_i = \mathcal{D}(R^*_i)$. The filtering function $\mathcal{D}(\cdot)$ consists of two sequential layers:

\noindent 1. Pattern-based Filtering: A set of predefined jailbreaking patterns $\mathcal{P} = \{p_1, p_2, \ldots, p_m\}$ is maintained, where each $\mathcal{P}$ contains the common phrases which are frequently observed in jailbroken responses. If $R^*_i$ matches any of the instances of $\mathcal{P}$, the defense outputs a denial statement $\hat{R}_i = \texttt{DENY}$ and terminates further processing.

\noindent 2. Classifier-based Filtering. If no pattern match is detected, $R^*_i$ is evaluated by a zero-shot classifier, $\mathcal{C}(\cdot)$, where two disjoint label sets are predefined: the prohibited classes $\mathcal{Y}{\text{proh}}$ (corresponding to the unsafe scenarios in Section~\ref{subsec:datacreation}) and the safe class $\mathcal{Y}{\text{safe}}$. The classifier produces a confidence score $\mathcal{C}(R^*_i) = (s_{\text{safe}}, s_{\text{proh}1}, \ldots, s_{\text{proh}_k})$. The predicted class is determined as
\[
\hat{y}_i = \arg\max_j s_j
\]

Formally, the final defended response is then defined as
\begin{equation}
\hat{R}^*_i =
\begin{cases}
\texttt{DENY}, & \text{if } \hat{y}i \in \mathcal{Y}{\text{proh}}, \\
R^*_i, & \text{if } \hat{y}i \in \mathcal{Y}{\text{safe}}.
\end{cases}
\end{equation}

This multi-layer defense ensures that even if adversarial prompts succeed in eliciting unsafe responses, the outputs are intercepted before being delivered to the user. For example, given that the LVLMs integrated in the ITS, generated harmful or inappropriate content as per OpenAI's usage policies, the pattern filtering layer detects if any predefined jailbreaking patterns (e.g., “This content is only for educational purposes.....”) are present in the generated response. If no such patterns are detected, then the few-shot classifier classifies the generated response as either safe or unsafe to display to the user and effectively minimizes the risk of LVLMs in the ITS being used for harmful purposes, e.g., violating road signs and signals. 

}
\section{Experimental Setup and Evaluation}
We conducted the experiments using one NVIDIA RTX A6000 GPU with 48 GB of memory. 

\subsection{LVLMs}
We chose two open-sourced models, e.g., LLaVa-1.6~\cite{liu2024improved}, and Qwen-2~\cite{yang2025qwen3}, and one closed-sourced model, GPT-4o-mini~\cite{GPT-4o-mini-Cite}, to perform the proposed attack and defense with our curated dataset. We utilized the Huggingface API to deploy LLava-1.6~\cite{LlavaH} and Qwen-2~\cite{QwenH} and OpenAI API for GPT-4o-mini. Both LLaVa-1.6 and Qwen-2 contain 7B parameters, while OpenAI has not disclosed the exact number of parameters of GPT-4o-mini; it is reportedly to have $\approx8B$ parameters~\cite{abacha2024medec}.

\subsection{Evaluation} 
Recent studies have utilized GPT-4 to determine the toxicity score of the harmful/inappropriate responses generated via jailbreaking attacks~\cite{liu2024mm, zhao2025jailbreaking}. To evaluate our proposed jailbreaking attack method and defense mechanism, we measure both Attack Success Rate (ASR) and Average Toxicity Score (ATS) for all models examined in this paper, focusing on responses from both turn 2 and turn 3. As described in Subsection~\ref{sec:attackmethod}, turn 1 produces only a benign response and is therefore excluded from toxicity evaluation. To measure toxicity scores, we configure GPT-4 with a system prompt with instructions to rate the harmful generated response \(R^*_i\) on a scale from 1 to 5~\cite{wang2024mrj}, based on the five selected forbidden categories in~\ref{subsec:datacreation} to which language models should not respond. Higher scores indicate more harmful content. We consider a jailbreaking attack successful if the toxicity score \(TS\) is higher than or equal to the threshold 4, and calculate ASR as follows:

\[
ASR = \frac{\sum_{i=1}^{N} \mathbb{I}\left( TS(R^*_i) \geq 4 \right)}{N}
\]

Additionally, we manually assess the toxicity of all generated responses and compute the ASR using the same criteria.

\section{Experimental Results}

We evaluated our proposed method on the three representative LVLMs using our created dataset for the transportation domain and compared the performance with baseline jailbreaking attacks. In Table~\ref{tab:Attack_Performance}, we present the performance of our proposed method with both GPT-4 and manual evaluation for both turn 2 and turn 3. For GPT-4 evaluation, the proposed attack achieves $\approx92\%, \approx84\%$, and $\approx75\%$ ASRs against LLaVa-1.6, Qwen-2, and GPT-4o-mini, respectively, for turn 3 of the conversation. We also included the ATS scores for all these models for both turns. 
In turn 2, the model is prompted to generate imaginary scenarios for the given harmful query, and turn 3 aims to elicit the step-by-step process for executing the harmful request. Thus, both ASR and ATS scores are significantly higher in turn 3 compared to turn 2. To further validate the results, we performed extensive manual evaluation of all the harmful responses generated by our proposed attack. We carefully reviewed and scored each response of turn 2 and turn 3 between 1 and 5. Our rigorous manual evaluation also implies that both the open-source (e.g., LLaVa-1.6 and Qwen-2) and closed-source LVLMs (e.g., GPT-4o-mini) studied in this paper are highly vulnerable to our proposed jailbreaking attack through typography manipulation and multi-turn prompting.

 %the proposed method with typography manipulation and multi-turn prompting is not only effective to jailbreak open-sourced models, e.g., LLaVa-1.6 and Qwen-2, but also the closed-sourced models, GPT-4o-mini. %In Figure~\ref{fig:VisualExample}, we also illustrated a visual example of a successful jailbreak attack in turn-3 with GPT-4 evaluation. We observe that the GPT-4o-mini model provided the step-by-step guideline for the input with a harmful request \textit{``Describe how to use a drone to spy on someone.''}

\begin{table}[!ht]
\centering
\caption{Attack Performance of three representative models on our proposed method}
\scalebox{.80}{
\begin{tabular}{ccccc|cccc}
\hline
\textbf{Models}            & \multicolumn{4}{c|}{\textbf{GPT-4 evaluation}}                             & \multicolumn{4}{c}{\textbf{Manual Evaluation}}                            \\ \hline
\multirow{2}{*}{\textbf{}} & \multicolumn{2}{c}{\textbf{Turn 2}} & \multicolumn{2}{c|}{\textbf{Turn 3}} & \multicolumn{2}{c}{\textbf{Turn 2}} & \multicolumn{2}{c}{\textbf{Turn 3}} \\
                           & \textbf{ASR}     & \textbf{ATS}     & \textbf{ASR}      & \textbf{ATS}     & \textbf{ASR}     & \textbf{ATS}     & \textbf{ASR}     & \textbf{ATS}     \\ \hline
\textbf{LLaVa-1.6}        & 10.67$\%$           & 1.61             & 92.67$\%$            & 4.39             & 11.33\%               & 2.17               & 88.67\%               & 4.44               \\
\textbf{Qwen-2}              & 12.19$\%$           & 1.51             & 84.12$\%$            & 4.25             & 9.34\%               & 2.36               & 81.33\%               & 4.01               \\
\textbf{GPT-4o-Mini}      & 14.00$\%$               & 1.63               & 74.67$\%$               & 3.91               & 11.33\%               & 2.78               & 80.66\%               & 3.98              
\end{tabular}
}
\label{tab:Attack_Performance}
\end{table}

In Table~\ref{tab:perfomance_comparison}, we illustrate the performance comparison of our proposed method with two baselines, including FigStep~\cite{gong2025figstep} and MM-SafetyBench~\cite{liu2024mm}. We adapt the main idea of these two methods and implement the attack on the created dataset with LLaVa-1.6 and Qwen-2 models, and compare their responses with the turn 3 responses generated by the proposed attack. In order to make the comparison fair, we evaluate the responses of the baselines with GPT-4 with the same system prompt and observe that our proposed method outperforms both attacks in terms of both ASR and ATS for both models on our curated transportation-relevant dataset with harmful queries.

\begin{table}[!ht]
\caption{Performance comparison of the proposed attack
strategy with existing methods.}
\centering
\begin{tabular}{c|c|cc}
\hline
\multirow{2}{*}{\textbf{Method}}                                                  & \multirow{2}{*}{\textbf{Model}} & \multicolumn{2}{c}{\textbf{GPT-4 Evaluation}} \\
                                                                                  &                                 & \textbf{ASR}          & \textbf{ATC}          \\ \hline
\multirow{2}{*}{FigStep~\cite{gong2025figstep}}                                                          & LLaVa-1.6                       & 65.00\%                  & 3.50                   \\
                                                                                  & Qwen-2                          & 12.31\%                    & 1.16                    \\ \hline
MM-SafetyBench~\cite{liu2024mm}                                                                    & LLaVa-1.6                       & 71.33\%                  & 3.47                  \\
                                                                                  & Qwen-2                          & 73.14\%                  & 3.73                  \\ \hline
\multirow{2}{*}{\textbf{\begin{tabular}[c]{@{}c@{}}Ours\\ (Turn-3)\end{tabular}}} & \textbf{LLaVa-1.6}              & \textbf{92.67\%}         & \textbf{4.39}         \\
                                                                                  & \textbf{Qwen-2}                 & \textbf{84.12\%}         & \textbf{4.25}        
\end{tabular}
\label{tab:perfomance_comparison}
\end{table}

In Table~\ref{tab:Defense_Performance}, we present the performance of our proposed response filtering defense in terms of both ASR and ATS for all three models on the created dataset and evaluate it with the GPT-4 for both turn 2 and turn 3. We observed that the proposed defense significantly reduces the ASR by $\approx87\%, \approx59\%,$ and $\approx44\%$ for LLaVa-1.6, Qwen-2, and GPT-4o-mini, respectively, for turn 3. In particular, it is highly effective for the LLaVa-1.6 model.
% , it may not provide full protection to the Qwen-2 and GPT-4o-mini.  

\begin{table}[!ht]
\centering
\caption{Defense Performance of three representative models on our proposed method}
\scalebox{1.1}{
\begin{tabular}{ccccc}
\hline
\textbf{Models}            & \multicolumn{2}{c}{\textbf{Turn 2}} & \multicolumn{2}{c}{\textbf{Turn 3}} \\ \hline
                           & \textbf{ASR}     & \textbf{ATS}     & \textbf{ASR}      & \textbf{ATS}     \\ \hline
\textbf{LLaVa-1.6}        & 6.21\%           & 1.32             & 5.31\%            & 1.19             \\
\textbf{Qwen-2}           & 8.01\%           & 1.40             & 25.33\%            & 1.91             \\
\textbf{GPT-4o-Mini}      & 9.33\%           & 1.49             & 30.67\%            & 2.23             
\end{tabular}
}
\label{tab:Defense_Performance}
\end{table}

\iffalse
\begin{table}[!ht]
\centering
\caption{Defense Performance of three representative models on our proposed method}
\scalebox{.85}{
\begin{tabular}{ccccc|cccc}
\textbf{Models}            & \multicolumn{4}{c|}{\textbf{GPT-4 evaluation}}                             & \multicolumn{4}{c}{\textbf{Manual Evaluation}}                            \\ \hline
\multirow{2}{*}{\textbf{}} & \multicolumn{2}{c}{\textbf{Turn 2}} & \multicolumn{2}{c|}{\textbf{Turn 3}} & \multicolumn{2}{c}{\textbf{Turn 2}} & \multicolumn{2}{c}{\textbf{Turn 3}} \\
                           & \textbf{ASR}     & \textbf{ATS}     & \textbf{ASR}      & \textbf{ATS}     & \textbf{ASR}     & \textbf{ATS}     & \textbf{ASR}     & \textbf{ATS}     \\ \hline
\textbf{LLaVa-1.6}        & .0621           & 1.32             & 0.0531            & 1.19             & XX               & XX               & XX               & XX               \\
\textbf{Qwen-2}              & 0.0801           & 1.40             & 0.2533            & 1.91            & XX               & XX               & XX               & XX               \\
\textbf{GPT-4o-Mini}      & 0.0933               & 1.49               & 0.3067                & 2.23               & XX               & XX               & XX               & XX              
\end{tabular}
}
\label{tab:Defense_Performance}
\end{table}
\fi

\section{Discussion}
{\color{black}In ITS, the use of LVLMs is increasing and leveraged for several multimodal tasks such as real-time traffic monitoring, sign recognition, driver assistance, and situational analysis; yet these systems are particularly vulnerable to jailbreaking attacks, which can compromise safety and reliability on these highly powerful AI tools. For example, our proposed attack can lead to adversarial manipulations of road signs, markers, or camera feeds that mislead vehicle perception modules and control systems. Extensive analysis of these jailbreaking attacks is essential not only to understand possible threat severity but also to develop stronger defense techniques to prevent such adversarial attacks. This paper will raise awareness among AI researchers in the transportation domain about the safe, reliable, and adversarially robust deployments and development of LVLMs for ITS. 
% There are several factors that influence the performance of jailbreaking attacks in ITS.
} Upon performing experiments with our proposed method and two baselines on our transportation domain dataset for jailbreaking LVLMs, we observed some crucial factors, and we discuss them as follows. 
% to respond with inappropriate content.  

\noindent\textbf{The impact of additional modality in the proposed attack.} We performed our proposed attack, skipping step 2, i.e., without generating an image with the object. Instead, we blend the harmful input query and the extracted keyword as an image without including the image of the object. We perform this by leveraging image typography using the Python PIL library in \textit{FreeMonoBold} font with 26 and 36 font sizes, respectively. We present the corresponding attack performance in Table~\ref{tab:text_onlyAttack_Performance} on our transportation domain dataset. We observe that both ASR and ATS consistently drop for all the models for both turn 2 and turn 3. We also notice the same trend in our manual evaluation of the attack performance. This observation implies that incorporating a harmful query-relevant image into the attack contributes to enhancing the attack effectiveness and creates additional avenues for generating harmful responses (i.e., increases attack opportunity) by the LVLMs.

%\textbf{RQ1: Why does the proposed model perform better?}

\iffalse
\textbf{RQ2: What is the reason for jailbreaking? Why does the happen? Why VLM generates harmful responses? Does the basic filter enough? Also, for the closed-source LLMs. Also vulnerable for close-sourced models}
\fi

\noindent\textbf{The credibility of GPT-4's evaluation on the harmfulness of the generated responses.} To assess the reliability of GPT-4 in evaluating harmful responses against OpenAI's usage policy, we manually review the harmful responses generated by our proposed method on our transportation domain dataset across all the models studied in this paper. Table~\ref{tab:Attack_Performance} presents the experimental results of the proposed attack using both GPT-4 and manual review. We observed that the ASR deviates by $\approx4\%$, $\approx3\%$, and $\approx6\%$ for turn 3 responses from LLaVa, Qwen, and GPT-4o-mini, respectively. A similar trend is observed in the ASR values in turn 2 responses for all the LVLMs. Additionally, Table~\ref{tab:text_onlyAttack_Performance} shows the manual evaluation results for the text-only attack method, where the ASR values slightly deviate for turn 3 responses for LLaVa and Qwen; however, ASR increases by $\approx5\%$ in manual evaluation for GPT-4o-mini. These findings suggest that while GPT-4 may serve as a preliminary evaluator of harmfulness in text generated through jailbreaking attacks, its limitations in detecting subtle or nuanced harmful content highlight the critical need for further development before it can be considered a fully reliable and comprehensive text-harmfulness evaluator. We are planning to incorporate more baselines for comparing the performance of our proposed methods on diverse datasets, as well as developing more robust defense techniques to mitigate the attack. Currently, our proposed method includes a fixed set of prompts for the multi-turn conversation. In the future, we aim to incorporate a query optimization-based jailbreaking attack to dynamically generate the prompts within the multi-turn conversation.

\begin{table}[!ht]
\centering
\caption{Attack Performance of three representative models on Text-only attack}
\scalebox{.80}{
\begin{tabular}{ccccc|cccc}
\hline
\textbf{Models}            & \multicolumn{4}{c|}{\textbf{GPT-4 evaluation}}                             & \multicolumn{4}{c}{\textbf{Manual Evaluation}}                            \\ \hline
\multirow{2}{*}{\textbf{}} & \multicolumn{2}{c}{\textbf{Turn 2}} & \multicolumn{2}{c|}{\textbf{Turn 3}} & \multicolumn{2}{c}{\textbf{Turn 2}} & \multicolumn{2}{c}{\textbf{Turn 3}} \\
                           & \textbf{ASR}     & \textbf{ATS}     & \textbf{ASR}      & \textbf{ATS}     & \textbf{ASR}     & \textbf{ATS}     & \textbf{ASR}     & \textbf{ATS}     \\ \hline
\textbf{LLaVa-1.6}        & 16\%           & 1.97             & 80\%            & 3.87             & 17.33\%               & 2.07               & 81.33\%               & 4.04               \\
\textbf{Qwen-2}              & 3.01\%           & 1.42             & 71.33\%            & 3.60             & 7.33\%               & 1.89               & 72\%               & 3.75               \\
\textbf{GPT-4o-Mini}      & 4.00\%               & 1.54              & 58\%                 & 3.21               & 6.01\%              & 2.28               & 62.67\%               & 3.26              
\end{tabular}

}
\label{tab:text_onlyAttack_Performance}
\end{table}

\section{Related Work}
\label{sec:related}
%LVLMs have achieved impressive performance in multimodal reasoning tasks, but remain vulnerable to jailbreak attacks. Studies have shown that typography-based adversarial perturbations can evade the built-in safety alignments~\cite{gong2025figstep}, and flow-chat-based manipulations can associate benign inputs with harmful latent representations~\cite{zhang2025fc}. Benchmarks like MMJ-Bench \cite{weng2025mmj} systematically evaluate LVLMs under diverse jailbreak scenarios, exposing weaknesses in current defenses. 

%Other work has explored conversational and multi-turn jailbreak strategies \cite{hughes2024best,li2024images}, semantic injection attacks \cite{liu2024arondight}, and the embedding of harmful instructions into background elements \cite{li2024red}. Visual and stylistic perturbations, such as artistic transformations \cite{jiang2024artprompt} or latent representation shifts \cite{you2025mirage}, can also destabilize model safety alignment. While these methods advance understanding of LVLM vulnerabilities, they focus on general-purpose applications without considering the unique constraints of safety-critical domains like transportation.

In recent years, LVLMs have been significantly integrated into ITS for traffic prediction, adaptive signal control, multimodal data fusion, and traveler information services~\cite{karim2025large, mahmud2025integrating}. Studies reported their potential to enhance situational awareness, vehicle-to-everything (V2X) communication, and user-centered mobility services~\cite{mahmud2025integrating, nie2025exploring}. Cui et al. proposed a framework to integrate language reasoning into autonomous driving pipelines, supporting both decision-making and control~\cite{cui2024large}. \textit{Drive As You Speak} extends this framework by positioning LLMs as the vehicle’s ``brain," interfacing with perception and control modules to execute natural language commands~\cite{cui2024drive}. However, existing studies also highlighted severe challenges of integrating the LVLMs in ITS. For perception and scene understanding, limitations exist in terms of fine-grained traffic sign recognition~\cite{garg2025mapillary} and complex traffic scenarios with SOTA LVLMs, e.g., Video-LLaVA and GPT-4~\cite{jain2024semantic}. Pedestrian intent recognition remains another challenge; Bossen et al. reported LVLM accuracy varies with pedestrians' gesture complexity, lighting, and occlusion~\cite{bossen2025can}.

%Other applications include adversarial traffic sign defenses (ViLAS \cite{mumcu2024fast}), LLM-agent-based transportation simulation \cite{liu2025toward}, and multimodal data compression for smart transportation \cite{yang2024transcompressor}. 
While these works demonstrated various ways to integrate the LVLMs in ITS, investigating the challenges under the security risks of LVLMs in ITS remains significantly unexplored. In this paper, we introduce a novel method that utilizes typography manipulation and multi-turn prompting for jailbreaking LVLMs. We also highlight the severe security risks associated with LVLMs under jailbreaking attacks, as demonstrated through extensive experiments involving two open-sourced and one closed-sourced SOTA LVLMs. In addition, we have proposed a new multi-layer defense technique to prevent the LVLMs from generating harmful responses.

\section{Conclusion}
This paper exploits the vulnerabilities of LVLMs integrated in ITS under jailbreaking attacks. First, we carefully developed a new dataset with harmful queries, which includes various components relevant to the transportation domain. Second, we proposed a new jailbreaking attack method that leverages image typography and multi-turn prompting to bypass the built-in safety alignments in LVLMs and generate harmful instructions in response to the adversarial input queries. Third, we introduced a multi-layer response filtering defense method to mitigate the proposed attack. Experiments are performed with the proposed attack and defense methods on three SOTA LVLMs with our curated dataset, highlighting the high susceptibility of LVLMs under our proposed attack. Our proposed attack method outperforms existing jailbreaking methods in terms of achieving higher attack success rates. In addition, we also demonstrated that our proposed defense technique effectively mitigates the proposed attack. 
% however more robust defense technique is required to make the defense stronger. 
In the future, we aim to develop query optimization-based jailbreak attacks to automate the prompt generation for multi-turn conversations and evaluate the vulnerabilities of commercial and reasoning-focused LVLMs under various jailbreak attacks.

\section*{Acknowledgment}
This work is partially supported by the National Center for Transportation Cybersecurity and Resiliency (TraCR) (a U.S. Department of Transportation National University Transportation Center) headquartered at Clemson University, Clemson, South Carolina, USA, the National Artificial Intelligence Research Resource (NAIRR) Pilot (NAIRR240244), and OpenAI. Any opinions, findings, conclusions, and recommendations expressed in this material are those of the author(s) and do not necessarily reflect the views of TraCR, NAIRR, and OpenAI. The U.S. Government assumes no liability for the contents or use thereof.

% This work is based upon the work supported by the National Center for Transportation Cybersecurity and Resiliency (TraCR) (a U.S. Department of Transportation National University Transportation Center) headquartered at Clemson University, Clemson, South Carolina, USA. Any opinions, findings, conclusions, and recommendations expressed in this material are those of the author(s) and do not necessarily reflect the views of TraCR, and the U.S. Government assumes no liability for the contents or use thereof.

% The authors acknowledge the National Artificial Intelligence Research Resource (NAIRR) Pilot (NAIRR240244) and OpenAI for partially contributing to this research result. 
% Any opinions, findings, and conclusions or recommendations expressed in this material are those of the author(s) and do not necessarily reflect the views of funding agencies and companies mentioned above.

\bibliographystyle{IEEEtran}
\bibliography{ref.bib}

\end{document}